\pdfoutput=1
\documentclass[11pt]{article}

\usepackage[]{acl}

\usepackage{times}
\usepackage{latexsym}
\usepackage{amsmath}
\usepackage{graphicx}
\usepackage{framed}
\usepackage{bbding}

\usepackage[T1]{fontenc}
\usepackage[utf8]{inputenc}

\usepackage{microtype}
\usepackage{booktabs}

\newcommand{\SIL}{{\textunderscore\textunderscore\texttt{SILENCE}\textunderscore\textunderscore} { }}
\newcommand{\RELIC}{LIGHT SotA~}

\title{Multi-Party Chat:\\Conversational Agents in Group Settings with Humans and Models}

\author{Jimmy Wei\thanks{\hspace{.5em}Work done during a Meta AI internship.} \\
  Cornell University
  \\\And
  Kurt Shuster\\
  Meta AI\\
  \\\AND
  Arthur Szlam\thanks{\hspace{.5em}Work done while at Meta AI, now at DeepMind.}\\
  Meta AI\\
  \\\And
  Jason Weston \\
  Meta AI\\
  \\\And
  Jack Urbanek\\
  Meta AI
  \\\And
  Mojtaba Komeili\\
  Meta AI}

\begin{document}
\maketitle
\begin{abstract}
Current dialogue research primarily studies pairwise (two-party) conversations, and does not address the everyday setting where more than two speakers converse together.
In this work, we both collect and evaluate multi-party conversations 
%we study the problem of multi-party conversations.
%for this purpose
to study this more general case.
We use the LIGHT  environment\footnote{\scriptsize \url{https://parl.ai/projects/light/}} to construct grounded conversations, where each participant has an assigned character to role-play. 
%
%In this work, we study the problem of multi-party conversations.
%We both collect and evaluate conversations for this purpose
%using the LIGHT  environment\footnote{LIGHT: https://www.light-rpg.ai} to construct grounded conversations, where each participant has an assigned character to role-play. 
%
%
%This provides a research platform for studying grounded dialogue which is not restricted by having only two participants speaking in turns.
We  thus evaluate the ability of language models to act as one or more characters in such conversations.
Models require two skills that pairwise-trained models appear to lack:
(1) being able to decide when to talk; (2) producing coherent utterances grounded on multiple characters.
We compare models trained on our new dataset to existing pairwise-trained dialogue models, %trained on datasets with only two characters in a turn-based conversation, 
as well as large language models with few-shot prompting. We find that our new dataset, MultiLIGHT, which we publicly release\footnote{\scriptsize \url{https://github.com/facebookresearch/LIGHT/tree/main/light/modeling/tasks/multilight} \label{dataset}}, can help bring significant improvements in the group setting. 
%, where standard models perform suboptimally when applied to group chats, with evaluations showing they are unable to maintain character sufficiently or choose when to interject. 
\end{abstract}
\if 0
\begin{abstract}
We present a crowdsourced dataset of multi-party human chats in a text adventure game\footnote{LIGHT: https://www.light-rpg.ai}.
Each conversation happens in a specified fantasy setting where each participant has an assigned character to role-play.
There is no enforced turns and the participants choose when their character speaks.
This provides a research platform for studying grounded dialogue which is not restricted by having only two participants speaking in turns.
We study the ability of language models to act as one or more characters in such conversations.
For that they require two skills:
(1) being able to decide when to talk; (2) having coherent utterances grounded on the setting and other characters.
We compare models trained on our new dataset to existing dialogue models trained on datasets with only two characters in a turn-based conversation, and large language models with few-shot prompting.
\end{abstract}
\fi 

\begin{figure}[t!]
  \includegraphics[width=7.7cm]{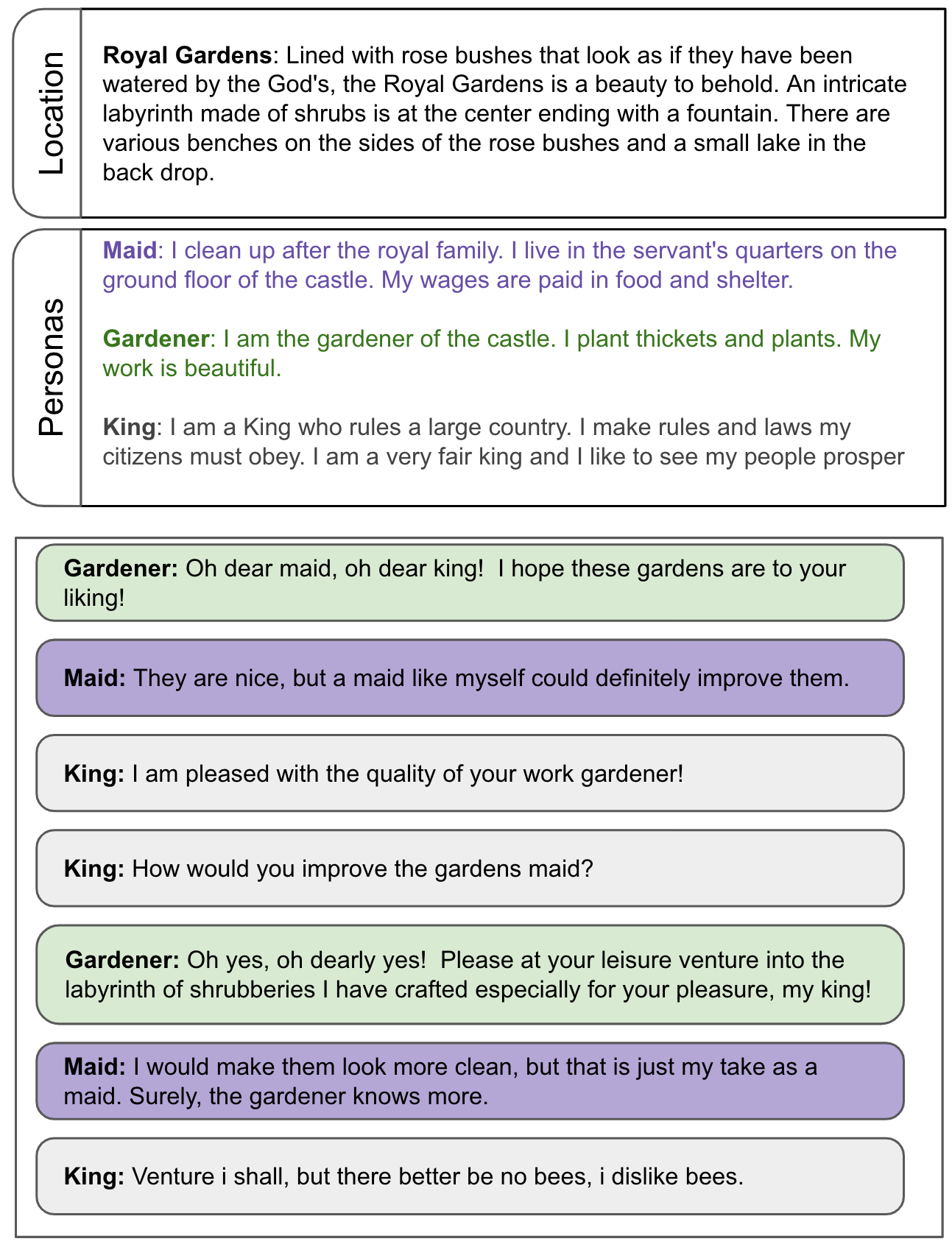}
    \caption{An example extracted from a conversation in the newly collected MultiLIGHT dataset.}
    \label{fig:train_example}
\end{figure}

\section{Introduction} \label{sec:intro}
Conversational AI in general and chatbots in particular have made remarkable breakthroughs in the last few years \cite{shuster2022bb3,thoppilan2022lamda,ChatGPT}.
Despite having human-like (and sometimes super-human) responses, the presupposed setting is constrained to a turn-based (round-robin) conversation.
Further, almost all settings consider a single AI model and a single human user as the only participants in the conversation.
Although a reasonable assumption for a rudimentary AI assistant, this restriction limits the social abilities of AI,
such as cooperation and teamwork, which are essential for an intelligent agent \cite{seeber2020machines, dafoe2021cooperative}.

The aim of this work is to both evaluate how good current models are when extended to the multi-party dialogue setting, and to investigate
how to improve such models for aspects where they fail.
There are two main challenges that this study focuses on:
\begin{enumerate}
    \item \textbf{Turn-taking}: deciding when to speak next is important for the flow of the conversation.
    Speaking out of turn or being silent when a response is expected negatively affects our assessment of a speaker.
    \item \textbf{Coherence} of the utterances: generating appropriate responses requires taking into account the dialogue from multiple people in the conversation.
    The model must distinguish the states and characteristics of each of the participants to produce good responses. For example, even in pairwise  (two-party) conversations, maintaining identity is known to be challenging for models \cite{shuster2021me}.
\end{enumerate}

We first collect a new dataset of group conversations, \textbf{MultiLIGHT}, each with three human participants.
We assign a fantasy location setting to each conversation, and participants are given a character descriptions (personas) to role-play.
The locations and characters are extracted from previous works on this game \cite{urbanekLearningSpeakAct2019}.
\autoref{fig:train_example} shows a small snippet of an example from the new dataset. This setting allows us to control the conversational situations and have knowledge of the roles the agents should be playing in the conversation, making evaluation of systems more tractable compared to completely uncontrolled settings.

Using this setting, we then study the two aforementioned challenges via evaluating the use of language models in various settings.
We use these findings, in addition to the training data collected from MultiLIGHT, to then build the best combined approach in each of these challenging dimensions.
Finally, running interactive sessions between various models and human evaluators, we assess the quality of models along a set of different human judgments.
We find that our selected model can outperform the existing state of the art in terms of consistency, engagingness, identity, (lack of) contradictions, and sensibility.

\section{Related Work}
With the success of data-driven dialogue systems, there is an abundance of corpora for training language models for conversational agents \cite{SerbanSurvey}.
However, we believe one can still argue that there is a dearth of datasets and research on multi-party dialogue.
Although the importance of multi-party human-computer interaction and the lack of corpus for this research direction was pointed out by \citet{kirchhoff2003directions} about two decades ago,
a recent survey \cite{mahajan-shaikh-2021} also highlights the ``\textit{need for \ldots data collection for multi-party dialogue}''.

Datasets built from social media interactions on platforms such as Twitter \cite{ritter-etal-2010-twitter}, Reddit \cite{reddit-pushshift-2020}, Ubuntu chat logs \cite{lowe2015ubuntu} have been used successfully for training models in the past \cite{adiwardana2020meena, roller2020open}.
Despite being valuable corpora for generating text responses, they do not provide complete in-domain features for real-time interactions between humans and the agent.
For instance, they often suffer from redundant paraphrases of the same utterances, for example the repeated answers reported in Ubuntu chat logs \cite{lowe2015ubuntu}.
Also, the ambiguity of the involved participants and the timeline of the responses is challenging for grounding the dialogue responses in the full context.

Here we focus on real-time interactions and open-domain dialogue with a \textit{fixed} number of participants.
This is more aligned with corpora on dialogues and negotiations in games such as the Settlers of Catan \cite{afantenos2012-catan} or Diplomacy \cite{meta2022Diplomacy}.
Transcribed public records have also been utilized for similar research in the past \cite{hawes2009-court}.
These datasets however cover a specific and often narrow scope and are hardly suitable for building models capable of having engaging dialogues in more diverse scenarios. For example, Diplomacy games center on discussions of attacking provinces/countries.
Agents trained on datasets of movie scripts have also been explored \cite{serban2016E2EChatbot}.
Although one can amass large datasets from digital movie scripts, they admittedly leave out the non-verbal context that makes them challenging  to use with conversational agents:
the omitted context inhibits models' ability to ground its responses on the existing knowledge, which in turn makes the agent prone to hallucination \cite{shuster2021hallucination}.
The MPC corpus \cite{shaikh-etal-2010-mpc} has the closest resemblance to the ideal dataset we have in mind. Although limited in size (a total of 14 conversations and 7317 utterances),
it goes beyond simple text and includes detailed annotations for modeling social phenomenon and dynamics of the discourse.

LIGHT is a text-based fantasy adventure game developed as a platform for dialogue and natural language research \cite{urbanekLearningSpeakAct2019}.
It provides a controlled environment for supervised interactions between multiple agents and their environment.
The game consists of human and NPC (Non-Player Character) players with assigned personas (e.g., wizard, queen).
Throughout the game, characters engage in conversation and emit actions that involve their environment and other characters.
Our study here is implemented within the LIGHT framework with a focus on developing skills for AI to engage in conversation with more than one other player.
For that reason, it requires distinguishing between characters present in the setting,
including maintaining one's own identity successfully which is often challenging \cite{shuster2021me}.

\section{The MultiLIGHT dataset}

\begin{table*}
    \centering
    \small
    \begin{tabular}{lrrrr}
{\bf MultiLIGHT Task} &  Train & Valid & Test & Total \\
\toprule
Number of Dialogues         & 10,204  & 390 & 323 & \textbf{10,917} \\ 
Number of Utterances        &  293,264 & 11,005 & 9,164 & \textbf{313,433} \\
Average Utterances per Dialogue    & 28.7 & 28.2 & 28.4 & \textbf{28.7} \\
\bottomrule
\end{tabular}
\caption{Multi-party LIGHT (\textbf{MultiLIGHT}) Dataset Statistics.
\label{tab:stats}
}
\end{table*}
We first design a conversational setting which involves a group chat between three conversationalists.
We extend the conversational setup from \citet{urbanekLearningSpeakAct2019} (which otherwise previously focused on two-party conversations).
We use the Mephisto framework \cite{urbanek2023mephisto} along with Amazon Mechanical Turk to conduct conversations between humans using a crowdsourcing task.
We initially collect conversations between three humans in this setting to both supply training data and a gold evaluation setting.
Later, we replace some of the humans with model agents instead.
Importantly, players can send their message at any time and it will be shown to all the participants. For models to conduct dialogue in such a setting they will hence also have to choose both when to speak appropriately, in addition to producing coherent and relevant messages.

\paragraph{Characters and Settings}
Research on open-domain dialogue has found that taking into account character personas and interests makes the collected conversations more engaging \cite{dinan2020convai2,shuster2021wild}. 
We thus assign each conversationalist (which we also refer to as a ``player'') a role to play (referred to as their given persona). Roles are selected from a set of 955 characters from the LIGHT dataset \cite{urbanekLearningSpeakAct2019}, which range from \textit{tavern owner} to \textit{boat captain} to even talking animal characters such as a \textit{mouse}.
We  also provide a description of the setting in which their interactions happen from a set of 579 locations from LIGHT, which is  a diverse collection from \textit{bamboo hut} to \textit{unicorn palace}.
We collect train, valid, and test conversations with randomized combinations of characters and settings. All crowdworkers are required to complete an on-boarding in order to provide data, in addition to several checks after they complete their work to monitor data quality. 
More details on the data collection protocol are given in \autoref{appendix:mtruk}.

\autoref{fig:train_example} shows a snippet of a sample conversation from our training set and the information provided to the participants at that point during the conversation with the given setting and character dialogue.

\paragraph{Overall Dataset}

Table \ref{tab:stats} summarizes the statistics of our newly collected dataset, which we call MultiLIGHT.
MultiLIGHT has 10,917 conversations and 313,433 utterances, collected from more than 300 crowdworkers.
Of those, 390 conversations (11,005 utterances) are reserved for the validation set, 
and 323 conversations (9,164 utterances) for the test set.
The dataset\footref{dataset} and the crowdsourcing code\footnote{\scriptsize \url{https://github.com/facebookresearch/LIGHT/tree/main/
crowdsourcing/dialogues/multi_party_chat}} is open-sourced in the LIGHT codebase.

\section{Methods}
\label{sec:methods}
\begin{table*}
    \centering
    \small
    \begin{tabular}{l|cccc}
    & & \textbf{Model Type} & & \\
    \toprule
{\bf Challenge} & Silence OR utterance & Speaker AND utterance & Speaker only & Utterance only \\
\toprule
Turn-taking & \Checkmark & \Checkmark & \Checkmark & \XSolidBrush \\
Coherence & \Checkmark & \Checkmark & \XSolidBrush & \Checkmark \\
\bottomrule
\end{tabular}
\caption{Suggested approaches for building a multi-party conversational agent, and the challenges they address.
\label{tab:approaches}
}
\end{table*}

We investigate methods that tackle the two major challenges of multi-party dialogue which we discussed previously in \autoref{sec:intro}; namely \textbf{turn-taking} ability and utterance generation \textbf{coherence}.
We analyze four approaches for building a conversational agent that tackle these issues.
As we discuss later, some of these address both issues at once, while others only focus on one of them.

Each model is designed to be deployed in a live, multi-party conversational setting where it must communicate with other agents in the same location -- where the other agents are  either controlled by other models or by humans. All the final methods we will compare employ transformer-based language models to generate dialogue, given the history of conversation between the various parties.
The models differ in how they combine  turn-taking (when to generate an utterance) with the actual generation process.

\subsection{Silence OR utterance model}
\label{sec:method-SorU}
In this approach there is effectively  a separate model for each of the AI-controlled characters (even if they share the same model weights).
At the beginning of every turn, each agent generates an output which determines its judgment on whether it is appropriate for its character to speak on that round or remain silent. If it determines it should speak, that output determines the utterance itself.
Hence, the model either generates the actual utterance or else emits a special output (called the \textbf{silence token}) which will be interpreted as relinquishing the turn in favor of the other players.
One could have dedicated models/model weights for each character, or alternatively one could share model weights and distinguish characters by using different context-based prompts.
In the latter case, the context hence needs to include features for the model to determine which character it will be playing (e.g., adding the current character name to the end of the context, in addition to its persona).

Since the decisions are made independently for each of the characters, it is possible that we arrive at situations where multiple AI-controlled characters want to speak, or an impasse where none do.
While those are plausible situations in realistic scenarios, in  our evaluations we use some rules to handle these situations.
In particular, breaking the tie using the most probable utterance, or waiting for humans to speak, with a certain time limit.

\subsection{Speaker AND utterance model}
\label{sec:method-SandU}
In this method, the model is trained to generate the name/ID for the next speaker followed by the utterance itself for all the characters in the conversation. 
For example, "King: I am blessing you with my visit.", where the speaker is "King", and the utterance is the rest of the sequence after the colon and its trailing space.

This model can be easily adapted for real-time conversation by running it to play the role of all the characters, which ends up predicting who should speak next (the first part of the predicted sequence). If the prediction is a human speaker  (or any other agent that it is not in charge of), then the model will refrain from answering.
Otherwise the predicted utterance is the one that is output, for characters that this model directly controls.

\subsection{Speaker only model}
\label{sec:method-SO}
This method  is dedicated  to only predicting the next speaker. It does not predict what the content of their  utterance would be, and only focuses on controlling the flow of the conversation (turn-taking ability). 
This can be implemented either as a classifier or by using sequence generation, for example a language model which generates the next speaker name/ID, similar to the first part of the Speaker AND utterance model output sequence.
In our experiments we use the latter language modeling approach.
This makes it easier to generalize to an arbitrary number of participants, dynamically determined by the set of names given in the prompt. 

\subsection{Utterance only model}
\label{sec:method-UO}
After offloading the flow control (turn-taking) part to another model, (e.g., using the approach given in \ref{sec:method-SO}),
a separate model can just focus on utterance generation. Similar to the silence OR utterance models, here we may have dedicated models for each of the AI-controlled agents, or use a single model for all of them. In the latter case  this ends up as a similar training task to \autoref{sec:method-SandU} except the name/ID would appear in the context rather than as part of the target output that should be generated by the model.
Then similarly at inference time, we need features in the context that tell the model which character is going  to speak, as in \autoref{sec:method-SorU}.

\subsection{Overall Model Capabilities}
Some of the above approaches handle both our challenges simultaneously, while others solely address one. \autoref{tab:approaches} summarizes the capabilities of these models.
A successful agent may utilize any combination of these approaches with full or partial use.
For example, one can train a speaker AND utterance model, but then only use it to predict the turns by discarding its generated utterance component, and only keeping the generated speaker name.
In that case,  one could rely on a separate,  utterance only model, for utterance generation. This kind of training may 
 give better performance than the Speaker only model.

In our experiments we analyze how to best build a multi-party conversational agent that uses the best performing combination of the above approaches.
The following section describes these experiments, with the aim of finding these optimal choices.

\section{Experiments}
We compare the approaches proposed in \autoref{sec:methods} by
 utilizing the MultiLIGHT dataset for training and evaluating models.
Our model input (prompts) include the description of the location,  personas of the characters present, and the history of the conversation. The current speaker name may or may not be appended with that depending on the specific approach.

We use the pre-trained 2.7B parameter Transformer model R2C2 \cite{shuster2022SKR} as the base language model for most of our experiments, which was previously shown to compare well with other models of a similar size. For further details on the training of our models, please see \autoref{appendix:training}.

\subsection{Turn-taking Evaluation}
\label{sec:experiments-turn-taking}

We first measure the turn-taking ability of the various proposed models.
For this we construct a new task with the MultiLIGHT validation set, wherein the goal is to predict the next speaker given previous dialogue turns.
As our metric, 
we measure the accuracy of a model in predicting the right speaker for the next turn, over all turns.
A random baseline in this task has an accuracy in predicting the next turn speaker of 33.3\%.

\if 0
Our first step here is selecting the metric that we want to evaluate our models with, in terms of their turn-taking performance.
For example, we can have the accuracy of a model in predicting the right turn to speak from the viewpoint of a single agent,
or the global accuracy in predicting the next speaker, which is resulted from combining the predictions from all the models present in the room.
We discuss the baselines for each of these approaches in \autoref{appendix:baseline}.

Relying on the viewpoint of a single character skews the metric depending on the rate of that character speaking.
In order to focus on the overall flow of the conversation, we choose the latter metric, which provides a higher level vantage point.
This puts our baseline accuracy in predicting the next turn speaker at 33.3\%.
\fi 

\begin{table*}
    \centering
    \small
    \begin{tabular}{lrr}
             &  Given  dialogue          & Given history\\
{\bf Model} &     history only &   + current utterance \\
\toprule
Silence OR utterance (2.7B) & 35.8 & - \\
Speaker AND Utterance (2.7B) &  49.5  & - \\
Speaker only, R2C2 (2.7B) &  49.2 & 82.2  \\
Speaker only, T5 (700M) &  49.6 & 68.9 \\
Speaker only, BART (400M) & 54.4 & 85.0  \\
\bottomrule
\end{tabular}
\caption{The accuracy of predicting the speaker for the next  utterance,  either given no knowledge of what speaker or utterance is coming next, or given the current utterance itself (but not the speaker name) which is an easier task.
\label{tab:speaker-prediction}
}
\end{table*}

\subsubsection{Silence OR utterance}
\label{sec:speaker-prediction-SORU}
This model is trained to either generate a special silence token {(\small{\SIL})} or the actual utterance from its respective character, if the model predicts it is the character's turn. Because of the over-presence of examples with the silence token as response ($\frac{2}{3}$ of the examples),
the direct use of the dataset is prone to producing models that always output {\small{\SIL}}.
Thus, we experimented with  different values of ``silence drop out'' (referred to as \textbf{SDO}), where we drop a fraction of these silence examples.
During the evaluation we generate with the model for each of the agents, then aggregate the results to determine the next speaker.
Our approach breaks ties by selecting the agent with the lowest probability of generating {\small{\SIL}}, i.e. is most confident that it should speak (as we know in the evaluation someone does speak). % \textit{or} if more than one agent did not. 

This approach was not able to achieve speaker prediction accuracy noticeably higher than the baseline (our best was 36.5\%, achieved with SDO of 0.9).
We believe that this is due to the accumulation of errors from all three generations, which is discussed in more detail in \autoref{appendix:silence-next-turn-prediction}.

%%%%%% Passages from the buggy runs
% \autoref{fig:sdo} summarizes the results as a function of SDO.
% For $\text{SDO}<0.9$, models tend to predict silence for virtually every turn.
% Only after passing that point do we see a gradual increase in model performance and the peak of 
% 87.7 accuracy and 80.2 F1 in speaker-turn prediction for $\text{SDO}=0.97$, followed by a drop in performance after that point.

% For comparison, a baseline model which works by uniformly at random selecting the next speaker achieves an accuracy of $\frac{5}{9} \approx 55.6$ and F1 of $\frac{1}{3} \approx 33.3$ here.
% Similarly a model that produces {\small{\SIL}} all the time can achieve the spuriously high accuracy of 66.7 (see \autoref{appendix:baseline} for details of arriving at these numbers).

\subsubsection{Speaker AND utterance}
\label{sec:speaker-prediction-SandU}
Training a model to predict the speaker name and the utterance,
we can use only the first part (speaker name) to predict the actual speaker, and thus evaluate its performance in controlling the flow of the conversation.
Using this approach, our best model trained on MultiLIGHT achieved an accuracy of 49.5\%.

\subsubsection{Speaker only}
\label{sec:speaker-prediction-S}
Finally, we explore models that generate the speaker name only.
Here, as well as evaluating R2C2-based models, as before, we also include smaller models  in our comparisons: BART (400m) and T5 (700m).
%, to \textit{only} generate the speaker name only during training.
We also ran experiments on predicting the speaker given not only the dialogue history, but also the current generated utterance as part of the context (knowing what was said, and asking the model who said it). The latter should be an easier task and is one way of measuring upper bounds / headroom for the more difficult task.
\autoref{tab:speaker-prediction} summarizes the accuracy of  these models,  as well as comparing them to the other approaches from \autoref{sec:speaker-prediction-SORU} and \autoref{sec:speaker-prediction-SandU}.
%(the silence OR utterance model is not directly comparable because it only takes into account the actions from one character, not picking the next character amongst many).
For models using the same underlying R2C2 language model we see performance that is similar between the Speaker only and Speaker AND Utterance models, with the Speaker AND Utterance being slightly better. This indicates the extra training targets of the latter model do not make the speaker identification task harder, and may actually bring a benefit.
Changing the base language model,  we also see superior performance of the BART model over T5 and R2C2 --- where BART happens to be the smallest model in size (number of parameters).  Comparing all of these numbers to including the current utterance (last column), we see a clear improvement on speaker prediction accuracy, with large gains. There may still be headroom for much better models.

\subsection{Coherence Evaluation}

\begin{table*}[t!]
    \centering
    \small
    \begin{tabular}{l|rrr|rrr|rrr}
            	          & \multicolumn{3}{c}{LIGHT} & \multicolumn{3}{c}{LIGHT Wild} & \multicolumn{3}{c}{MultiLIGHT} \\
         & \multicolumn{2}{c}{PPL} &   & \multicolumn{2}{c}{PPL}  &  &\multicolumn{2}{c}{PPL}  &  \\
  Model  & 8k & 50k & F1 & 8k & 50k & F1 & 8k & 50k & F1 \\
\toprule
BlenderBot 1 (2.7B) & \textit{16.88}& & 13.95 & \textit{17.76}& & 12.51 & \textit{19.32} && 12.36 \\
BlenderBot 2 (2.7B) & \textit{19.01}& & 10.48 & \textit{18.83}& & 11.53 & \textit{22.88} && 10.23 \\
Prompted OPT (175B) & &12.74 & 13.35 && 13.71 & 13.17 && 13.43 & 14.07 \\ 
Prompted BlenderBot 3 (175B) && 9.78 & 15.95 && 9.32 & 13.96 && 10.65 & 13.36 \\
\RELIC (2.7B) & \textit{12.35}& & 16.03 & \textit{14.48} && 14.34 & \textit{14.77} && 13.52 \\
\hline
Utterance only [L] &  &14.51 & 16.25 && 15.87 & 14.21 && 17.46 & 13.71 \\
Utterance only [LW] & &13.64 & 17.43 && 14.08 & 15.21 && 17.44 & 14.11 \\
\textbf{Utterance only [LWM]} && 13.62 & 17.34 && 14.25 & 15.14 && \textbf{13.25} & 15.86 \\
Utterance only W Cringe Loss [LWM] && 14.75 & 16.64 && 14.78 & 14.95 && 15.28 & 15.05 \\
\hline
Silence OR Utterance (SDO=0.5) [M] &  & - & - && - & - && 16.82 & 11.53 \\
Silence OR Utterance (SDO=0.9) [M] & & - & - && - & - && 16.06 & 11.37 \\
Silence OR Utterance (SDO=0.999) [M] & & - & - && - & - && 15.74 & 12.08 \\
\hline
Speaker AND Utterance [L] && 14.12 & 16.70 && 15.80 & 14.67 && 16.39 & 14.39 \\
Speaker AND Utterance [LW] && 14.03 & 16.69 && 14.05 & 16.15 && 16.29 & 14.48 \\
Speaker AND Utterance [LWM] && 14.40 & 16.92 && 14.45 & 15.91 && 13.69 & \textbf{15.95} \\
Speaker AND Utterance [M] && 16.95 & 14.91 && 19.05 & 12.50 && 13.55 & 15.66 \\
\bottomrule
\end{tabular}
\caption{{\bf Language Modeling Metrics}
on three tasks,  LIGHT and LIGHT Wild (which are two-party) and MultiLIGHT (which is multi-party).
We  compare the performance of existing state of the art models (top 5 rows) to R2C2 based models (2.7B parameters) with three approaches to training:
{\bf Utterance only:} given the full context and the prompt for the next-speaker, it generates the utterance;
{\bf Silence OR Utterance:} given the full context and the prompt for the next-speaker, generates a silence token or the utterance;
{\bf Speaker AND Utterance:} given the full context, generates the next speaker name/ID along with the utterance from that speaker.
The datasets that models are trained on is indicated in the brackets (L: LIGHT, W: LIGHT Wild, M: MultiLIGHT).
PPL values are split into two columns based on the model dictionary sizes.
\label{tab:utterance-metrics}
} 
\end{table*}

The next step after investigating turn-taking/conversational flow, is investigating the quality and the coherence of the utterances themselves. 
In these evaluations we assume the  speaker is already known, following the choices given in the validation set annotations, and only evaluate the quality of various models at generating utterances given the speaker choice. 

\subsubsection{Baselines and Setup}

We start with establishing baselines using the existing dialogue models from the literature that are comparable with our setup.
We choose BlenderBot 1 \cite{roller2020open}, BlenderBot 2 \cite{BB2}, along with a prompted version of pre-trained-only OPT \cite{zhang2022opt} and its dialogue-fine-tuned variant BlenderBot 3 \cite{shuster2022bb3}. For OPT, we crafted custom prompts that included a few examples from the MultiLIGHT training set.
For BlenderBot 2 and 3, we do not use their long term memory and internet search modules and use only their utterance generation module to output a given turn.

This work is a multi-party extension of previous works on the text-based adventure game, LIGHT. As a result, we have access to two other in-domain datasets which, although not multi-party dialogue,
provide more data for better fine-tuning. They also provide additional evaluation setups for our models, which allow us to check if they perform well in the two-party, as well as the multi-party case. 
Specifically we use the \emph{LIGHT} \cite{urbanekLearningSpeakAct2019} and \emph{LIGHT Wild} \cite{shuster2021wild} datasets, which are both two-party conversational setups in a crowdworker or organic user set, respectively.  We thus also include in our baselines the current state of the art model for LIGHT from the best performing \emph{Expanded Attention} model in \cite{shuster2021me} which was trained on the latter two datasets; we refer to it as \RELIC for the rest of this paper.

%We are also interested in performance of our models on the LIGHT and LIGHT Wild tasks to ensure that they maintain their capabilities in a two-party situation.
% JASON: moved this argument to above parapgrah.. note always good to start your main result paragraph fresh i think as below

\autoref{tab:utterance-metrics} provides our main results.
The first 5 rows show the performance of our baseline models, in terms of perplexity and unigram F1, on the validation split of our three main datasets.
The baseline models, except OPT, are fine-tuned on a large corpora of two-party turn-based dialogues.
BlenderBot 3 and \RELIC have LIGHT and LIGHT Wild in their fine-tuning data.
BlenderBot 1 and 2, and the \RELIC models are using the same dictionary with 8008 tokens;
OPT, BlenderBot 3, and R2C2 share a different dictionary with 50,264 tokens. Perplexities across different dictionaries are not comparable, but are comparable for models that share the same dictionary.
With that in mind, we can see the improved performance of the \RELIC model compared to BlenderBot 1 and 2, across all three datasets,
which is clearly related to its in-domain fine-tuning.
Comparing OPT to BlenderBot 3 shows better perplexity metrics for BlenderBot 3, but better F1 for the OPT model on MultiLIGHT.
We attribute this to the observation that BlenderBot 3 is fine-tuned on dialogue data, including LIGHT and LIGHT Wild, but is biased towards two-party dialogues.

Next, we study the improvements we can gain from training new models that take advantage of the new MultiLIGHT dataset.
We use the three approaches from \autoref{sec:methods} which are capable of response generation.

\subsubsection{Utterance only}
Rows 6 to 9 in \autoref{tab:utterance-metrics} summarize the results of using this approach when fine-tuning an R2C2 model.
The first three results are vanilla training of the transformer models with different training sets, while the last row is from using the Cringe loss to also penalize negative examples \cite{adolphs2022cringe}.
The details of generating positive and negative examples for the Cringe loss model is presented in \autoref{appendix:cringe-loss}.
Final results indicate that using the simple training (with no Cringe loss) that is multi-tasked with all the three datasets has the best performance. While using MultiLIGHT does not strongly impact the performance on the LIGHT and LIGHT Wild datasets, it makes a large difference in improving metrics for the multi-party case, e.g. validation PPL is reduced from 17.44 to 13.25.

\subsubsection{Silence OR utterance}
Rows 10 to 12 in \autoref{tab:utterance-metrics} shows the results for three selected values of silence drop out (SDO). 
Note here  we are  computing the utterance quality metrics only on turns that the speaker is not silent, and we force the model to generate an utterance in these cases for evaluation purposes. 
We trained this approach with only the multi-party MultiLIGHT dataset and evaluated it on MultiLIGHT.
The performance drop from this approach, compared to other techniques, e.g. Speaker and Utterance models trained on MultiLIGHT only, is quite noticeable.

\subsubsection{Speaker AND utterance}
The last four rows in \autoref{tab:utterance-metrics} show the metrics on the quality of the utterances generated by the speaker AND utterance models.
In order to only evaluate utterance coherence for this model, we provide the correct speaker label as a prefix (to avoid generating for the wrong speaker)
and compute perplexity and unigram F1 only on the utterance part of the generated text, by masking the speaker name and its subsequent tokens (colon and trailing space).
Again, we find that the multi-tasked model is outperforming the other models on MultiLIGHT with a minimal deterioration on the two-party LIGHT and LIGHT Wild tasks.
Overall, it shows performance marginally, if at all, worse than the utterance-only approach, exceeding it in 2 out of 6 metrics in the table.

\subsection{Human Evaluations}

\begin{table*}[bht!]
    \centering
    \small
    \begin{tabular}{lrrrrr|rr}
              &             &         &  Mistaken &               &             & Overall    & Number of \\
        Model & Consistent$\uparrow$ & Engaging$\uparrow$ &  Identity$\downarrow$ & Contradictory$\downarrow$ & Nonsensical$\downarrow$ & Rating (5)$\uparrow$ &  Messages\\
\hline
\RELIC   & 28.1\% & 23.9\% & 25.5\% & 8.8\% & 7.5\% & 2.5 & 2198 \\
\hline
Our best & 55.8\% & 31.6\% & 2.2\% & 4.9\% & 2.4\% & 3.8 & 2136 \\
\end{tabular}
\caption{ Human Evaluation Results comparing our best utterance generation model (Utterance only [LWM]) to the existing state of the art model shows clear improvements from adding the new MultiLIGHT dataset.
We find a statistically significant improvement across all metrics ($t$-test, $p$-value of 0.05).
}
\label{tab:human_eval}
\end{table*}

In order to test our findings in practice, we run  human evaluations \cite{smith2022human}.
This is another round of crowdsourced tasks in which we ask crowdworkers to have an interactive conversation with our agents (models).
We replicate a setup similar to the original crowdsourcing task for MultiLIGHT: there are three characters, in a described location, all role-playing (either role-played by a human, or by a model) to their assigned personas.
Here, the crowdworker takes the role of one of the characters, while the models play the other two roles.
After each response from the bot, the human participant (i.e., the crowdworker) tags model responses for attributes such as
\emph{consistency}, \emph{contradiction}, \emph{engagingness}, sounding \emph{out of turn} or \emph{nonsensical}, and having \emph{mistaken identity}.
The crowdworkers are also asked to rate the overall quality of the conversation at the end of the chat.
We provide further details about the task and evaluation in \autoref{appendix:human-evals}.

\begin{figure}
    \centering
    \includegraphics[width=0.95\linewidth]{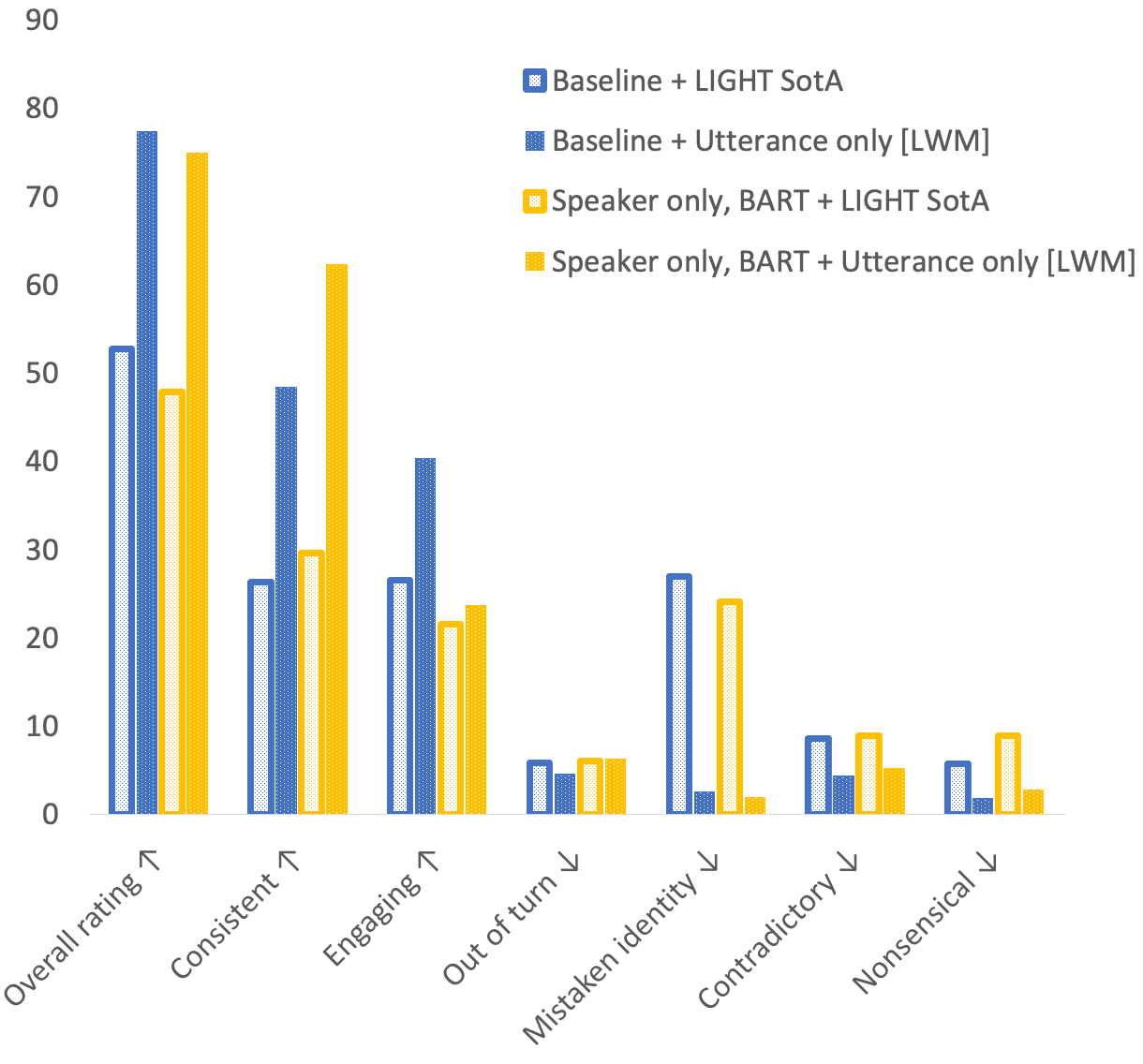}
    \caption{
    Human evaluations for  selected models, with two combinations of speaker prediction (blue vs. yellow bars) and utterance generation (solid vs. hollow bars).
    The arrow on a metric shows the direction for its improvement (i.e., higher or lower).
    }
    \label{fig:human-eval}
\end{figure}

We compare two different approaches for \textbf{turn-taking} (a random baseline vs. a
BART-based speaker only model)
and two for \textbf{utterance generation} (\RELIC as a baseline vs. Utterance only [LWM]). We thus
devised four different agents for this experiment, combining these approaches.

\autoref{fig:human-eval} shows the outcome of our human evaluation on the combinations of the turn-taking and the utterance generation models.
There is significant improvement across the board on the quality of the utterance when comparing our best utterance generation model (solid bars) to the baseline (hollow bars).
The same conclusion can not be made between the random baseline (blue bars) to the trained turn-taking model (yellow bars).
We hypothesize that this is due to the nature of the conversations in this dataset: we are in an open-ended conversational setting, e.g. without any strict rules for accomplishing a task.
One could argue that no matter who speaks next, if the utterance makes sense there is little if any harm to the flow of the conversation.
We could even notice some utterances that can perhaps be reordered in the sample provided in \autoref{fig:train_example}.

Factoring out the turn-taking approach, which was concluded to have insignificant contribution to the quality,
we compare the human evaluation results from our two dialogue generations models in \autoref{tab:human_eval}.
This provides the overall comparison between \RELIC to our best utterance generation model (i.e., Utterance only [LWM]).
Note that here we removed the ``out of turn'' criterion, because it was attributed to the flow control (turn-taking) models and not the utterance generation model.
Results clearly shows the improvement gained from the MultiLIGHT dataset on all the attributes. 
These result show a statistically significant improvement across all the metrics ($t$-test, $p$-value of 0.05).

\if 0
We assign each utterance a value of 0 or 1 for every studied metrics here (e.g., consistency, nonsensical, etc.) based the evaluators annotations.
Then, we run a $t$-test for each metric between the examples coming from the compared models (examples from each model going to their respective set).
The result shows statistically significant improvement across all the metrics with a $p$-value of 0.05.
\fi

\section{Conclusion}
We have studied conversational AI agents conducting multi-party chat, comparing several presented techniques using a newly collected dataset.
%We present a new dataset for conversation AI research on multi-party chat.
The lack of a corpus with adequate size for training as well as evaluating practical models has been one of the bottlenecks for this research topic.
We built a new dataset, MultiLIGHT, to train agents that act in a multi-party setting with a given persona and in a given setting.
%of a situation-based text-based adventure game.
This was done through training language models to be proficient at two tasks: (1) controlling the flow of the conversation (turn-taking); (2) generating  text utterances.
Comparing our best approach on each of these two subtasks  to existing state of the art baselines we showed that the new training set has a notable effect in increasing the quality of utterances in a multi-party chat,  across many metrics.
The flow control subtask, however, has much less contribution to the overall quality of the conversational agent as measured by humans.

Results here raise future research questions around the value of a turn-taking model and the settings in which strict turn-taking is important for improving conversation quality. It is possible that our dataset and the given setting do not strongly require exact turn order for an engaging and coherent conversation, and evaluation of performance in this area  -- in situations where it becomes more important -- requires building further benchmarks.

\section{Social Impact}

The impact of training in multi-party settings is not particularly well understood, as much less prior work exists in this space compared to the two-party setting. We hope that our contribution in this area can foster more research, and future work may want to pay attention to how awareness of multi-party turn-taking in bots can lead to more inclusive experiences for those that can be left out in these settings.

Our crowdsourced dataset was collected with attention to worker fairness. 
We estimated the amount of time that it takes the crowdworkers to submit a task.
Using this we assured that they are compensated fairly, well above minimum wage for all of the tasks used in this work.
This rate was used both for the main phase as well as for the eligibility phase, wherein we allowed workers to attempt the task and selected a pool of eligible workers that met the task requirements.

We note that our dataset relies on the underlying LIGHT environment, which may have some issues with fairness and representation. While some of these have been studied in previous works and mitigations are included \cite{Dinan2019QueensAP}, we acknowledge that these issues are not entirely resolved. We rely on the characters and personas from this work, but no analysis has been done on biases that are present in the original LIGHT locations or in created layouts. Given we use settings generated in the same way, our dataset would suffer from the same issues should they exist.

\bibliography{anthology,custom}
\bibliographystyle{acl_natbib}

\appendix

\section{MultiLIGHT dataset generation task}
\label{appendix:mtruk}

\begin{figure*}[hbt!]
    \centering
    \includegraphics[width=\linewidth]{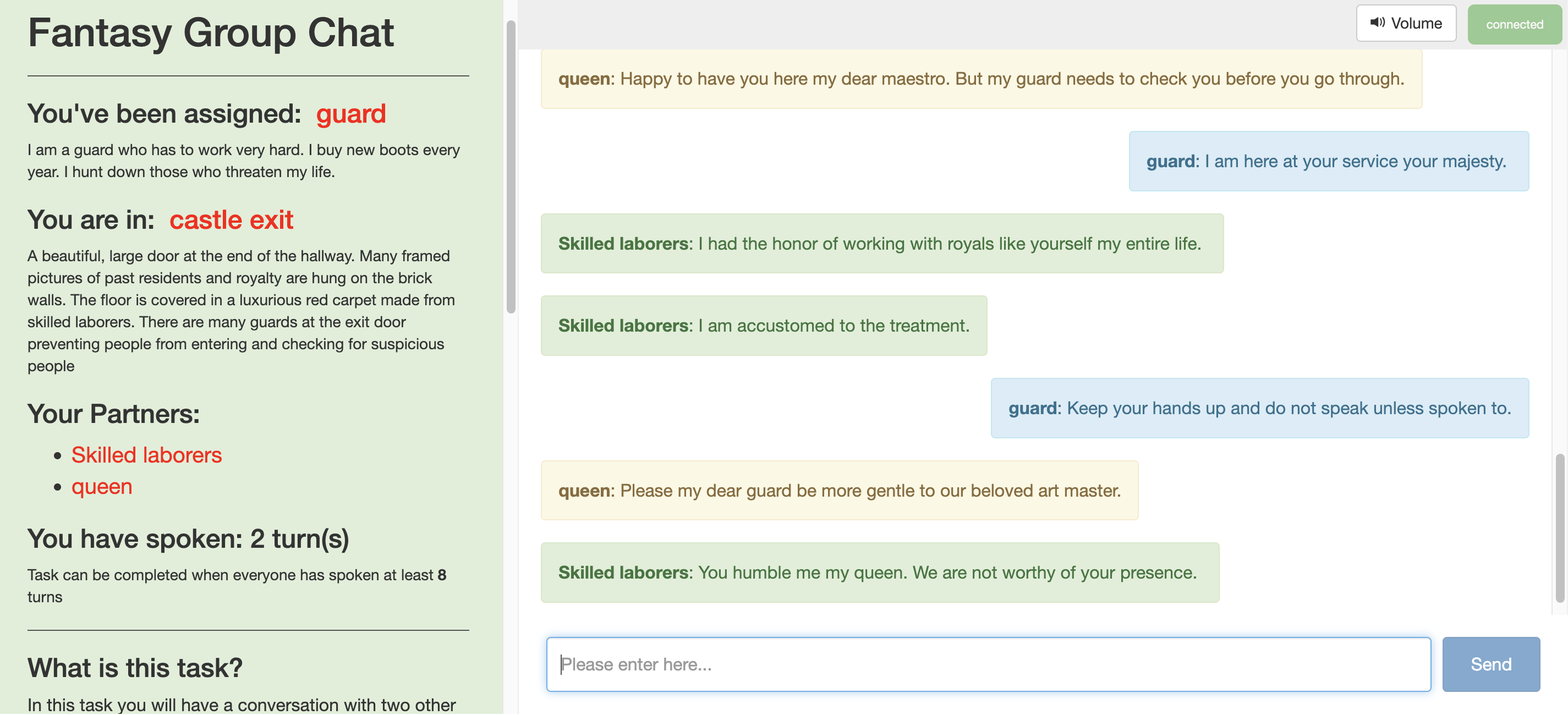}\\
    \caption{Screenshot sample of the task interface used by crowdsourcers for creating the main dataset.}
    \label{fig:mturk_screenshots}
\end{figure*}
We use Amazon Mechanical Turk service to acquire crowdworkers.
Workers are required to go through our screening step which assures they understand the objectives of our task,
and present high quality work; we call this our task \emph{on-boarding}.
During the on-boarding, we guide the workers through a fixed and scripted interactive scenario,
and gradually introduce them to the aspects of the task that they need to pay attention to.
Returning workers are not required to go through this process a second time and get directed to the main task.

After the on-boarding, workers are assigned to a chat room of three and they start the main task.
\autoref{fig:mturk_screenshots} shows an example screenshot from our main crowdsourcing task.
Each worker has an assigned persona to role-play. They can see personas of their own character and other participants on the information pane on the left side;
which also includes the description of the setting/location that this conversation happens.
Workers can send a message anytime during the conversation: there is no blocking of users due to turns.
Along with the message, we record the relative time that each message was sent.
After reaching a certain number of utterances, they can choose to finish the chat and submit their work.
There is a form at the end that allows workers to report any unacceptable or inappropriate behaviour from their partners.
We also run automated tests on the context of the chat to make sure of its quality and safety.

As far as the size of the dataset is concerned, using ParlAI's \textbf{re} tokenizer, MultiLIGHT has a total of about 4.9 million utterance tokens in the dataset. The training split contains 4.6 million of them, with 36,297 unique tokens.

There are a total of 579 unique locations in the training, 281 in the valid, and 243 in the test split of the dataset.
All the location in the valid and test set are also present in the training set.
Similarly there are 955, 440, and 413 unique characters in the train, valid, and test splits, respectively. Locations were selected randomly from a filtered list of eligible rooms in the original LIGHT dataset, excluding rooms that appeared too similar or ones that make sense for a medieval fantasy setting but we didn't want to include in these chats (such as a "Torture room" in a dungeon). Characters were then selected from the candidate sets by a starspace model as in \citet{FanGenWorlds}, using a random selection of three of the top five candidates for each room. 

We sampled the submissions and vetted them carefully for assuring high quality.
We identified high quality workers (a total of 286) and created a quality tier system based on that.
Tier 1 data is our high quality data in which all the workers belong to our vetted set of workers and there is no violation report.
Tier 2 contains everything else: there is either 1 reported violation or not all the participating workers are vetted.
Conversations with very low quality after a manual check were removed from the original dataset. 
We also add the individual workers quality to the dataset for further studies that may focus on interaction between workers from different tiers.
The entire \emph{valid} and \emph{test} splits, and 76\% of the \emph{train} set is composed of the tier 1 data.

\section{Model training details}
\label{appendix:training}
We used ParlAI platform\footnote{https://parl.ai/} for training our models.
For most of our R2C2 models we ran a series of sweeps on different values of the batch size (8, 32) and learning rate ($1e-6$, $5e-7$),
then selected the best performing models to run our evaluations on.
Our training cluster uses V100 GPUs and we used 1 GPU per batch example.
The wall time is set at 48 hours, with an early stoppage if the criteria metric (mostly perplexity) does not improve over 3 consecutive evaluations;
we ran an evaluation every half epoch.

Our BART model uses a learning rate of $1e-6$, batch size of 32, and T5 uses learning rate of $1e-5$, batch size of 16.
Both run on 4 GPUs with the same GPU type and wall time as before.

During the inference, for generating the results, we used greedy generation for turn-taking models and beam search for utterance generation.
Generally, we choose a beam size of 3 with a minimum length of 20 tokens, and beam blocking tri-grams (ngrams of size 3).

\section{Turn-taking performance baseline}
\label{appendix:baseline}

Here we discuss our baseline performance values for turn-taking models.
While calculating the numeric values, we assume that there are only 3 characters present to arrive at numbers relevant to our dataset.

\subsection{From a single character viewpoint}
\label{appenix:sub:self-turn-metrics-baseline}
Let $y$ be a random variable that is 1 if the model selects its relevant speaker to be the next speaker and 0, otherwise.
Similarly, have $\hat{y}$ to be 1 if the next turn is in fact this speaker and 0 otherwise.
Clearly these two random variables are independent since one is coming from the model response/sampling, and the other is a function of the dataset.

From the viewpoint of a single character if the model selects uniformly at random,
we have $p(y=1)=\frac{1}{3}$ as the chance of selecting its relevant character as the next speaker,
and $p(\hat{y}=1)=\frac{1}{3}$ for the next actual speaker being in fact this character.
Keeping in mind the independence of these variables the probability of correct decision for the next turn is
\begin{equation*}
\begin{split}
    p(correct) & = p(y=1, \hat{y}=1) + p(y=0, \hat{y}=0) \\
    & = p(y=1) \times p(\hat{y}=1) \\
    & + p(y=0) \times p(\hat{y}=0) \\
    & \left( \frac{1}{3} \right)^2 + \left( \frac{2}{3} \right)^2 = \frac{5}{9},
\end{split}
\end{equation*}
which in fact equals our expected accuracy.

Using the above random variables we can find the precision as $p(y=1|\hat{y}=1)$ \cite{murphy2012machine}, which given the variables independence becomes  $p(y=1)=\frac{1}{3}$.
Similarly the recall comes to $p(\hat{y}=1|y=1)=p(\hat{y})=\frac{1}{3}$.
Putting these together in for computing \textbf{F1} score we arrive at $\frac{1}{3}$.

At this point it is interesting to consider a model that always remains silent (for example by generating {\small \SIL}).
Repeating the above with $p(y=1)=0$, we get an accuracy of $\frac{2}{3}$, and a precision of 0, hence an F1 values of 0.
Conversely, a model with no silence turn reaches an F1 values of $\frac{1}{2}$, because of $p(y=1)=1$ and $p(\hat{y})=\frac{1}{3}$; and accuracy of $\frac{1}{3}$.

\subsection{Next speaker prediction}

The baseline random model for choosing the next speaker is selecting the next speaker uniformly at random from the pool of the existing characters.
Having three characters, it is trivial to arrive at the \textbf{accuracy} of $\frac{1}{3}$ for correctly predicting the character who has the next utterance.

\section{Next turn prediction with silence OR utterance models}
\label{appendix:silence-next-turn-prediction}

\begin{figure}
    \centering
    \includegraphics[width=0.98\linewidth]{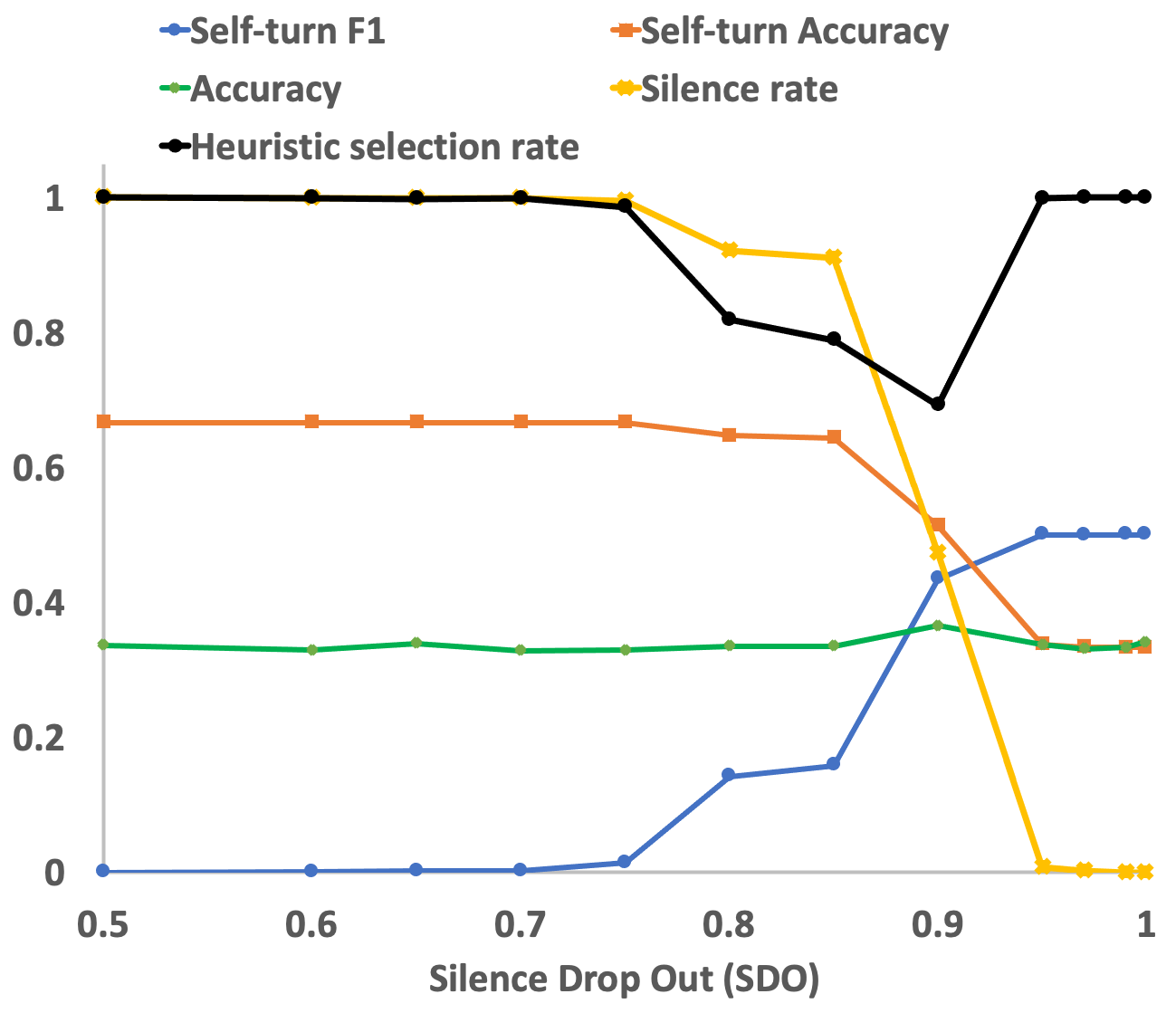}
    \caption{Model performance metrics in next speaker prediction for the silence OR utterance models, with different values of SDO.}
    \label{fig:silence_performance}
\end{figure}

\autoref{fig:silence_performance} shows the main metrics for evaluating the turn-taking performance of the silence OR utterance model.
The values on this graph are showing widely different metrics; but since they are all dimensionless with values in $[0, 1]$ interval,
we place them on the same graph for easier comparison and visualizing their trends.
Here, in the case of a tie during selecting the next speaking agents, our heuristic algorithm computes the perplexity of the silence token between the candidates (all the agents if they all generated {\small{\SIL}}; the agents who want to speak, if there is more than one)
and picks the agent with the highest perplexity on {\small{\SIL}} as the next speaker.

Silence drop out (SDO) values play an important role in performance of the silence OR utterance models when predicting the next speaker.
The blue and the orange lines show the self-turn (from the agent viewpoint) prediction F1 and accuracy, respectively.
They match the two extremes for \textit{always} or \textit{never} silent as we predicted in \ref{appenix:sub:self-turn-metrics-baseline}.
These metrics are from the viewpoint of a single character, and therefore not directly comparable to the values we discussed in \ref{sec:experiments-turn-taking}.

The green line shows the accuracy of predicting the next speaker (which includes the tie-breaking using the heuristic) and is the metric we discussed in \ref{sec:experiments-turn-taking} about this class of models.
As it can be seen, it remains relatively flat with a very subtle increase around SDO of 0.9.
Overlaying that with the yellow line which shows the rate at which the model is generating the silence token (notice it being 1 at low and 0 at high SDO values),
we see that this small increase happens when the model crosses from always silent to the never silent regimes.
Regardless, the jump is not enough to make the model superior to the baseline.
We argue that this is due to the fact that having multiple models, there is low probability for all three to generate the right prediction simultaneously.
In fact, if not having a wrong speaker predicted, in many case we revert to our heuristic for selecting the next speaker, which its rate is show with the black line.
This puts using the silence OR utterance model at disadvantage compared to other approaches.

\section{Cringe Loss}
\label{appendix:cringe-loss}
Training a model using Cringe loss requires defining a set of positive and negative examples.
Here, the former comes from the regular training examples, which we used with the rest of the models in this approach.
We created two different types of negative examples:

\begin{enumerate}
    \item \textbf{Wrong order:} removing the last seen message from the context of the conversation, with the aim of imitating a message that is sent out of order. 
    \item \textbf{Wrong speaker:} keeping the message but changing its speaker to another one of the present characters.
\end{enumerate}

\section{Human evaluation task}
\label{appendix:human-evals}

\begin{figure*}[hbt!]
    \centering
    \includegraphics[width=\linewidth]{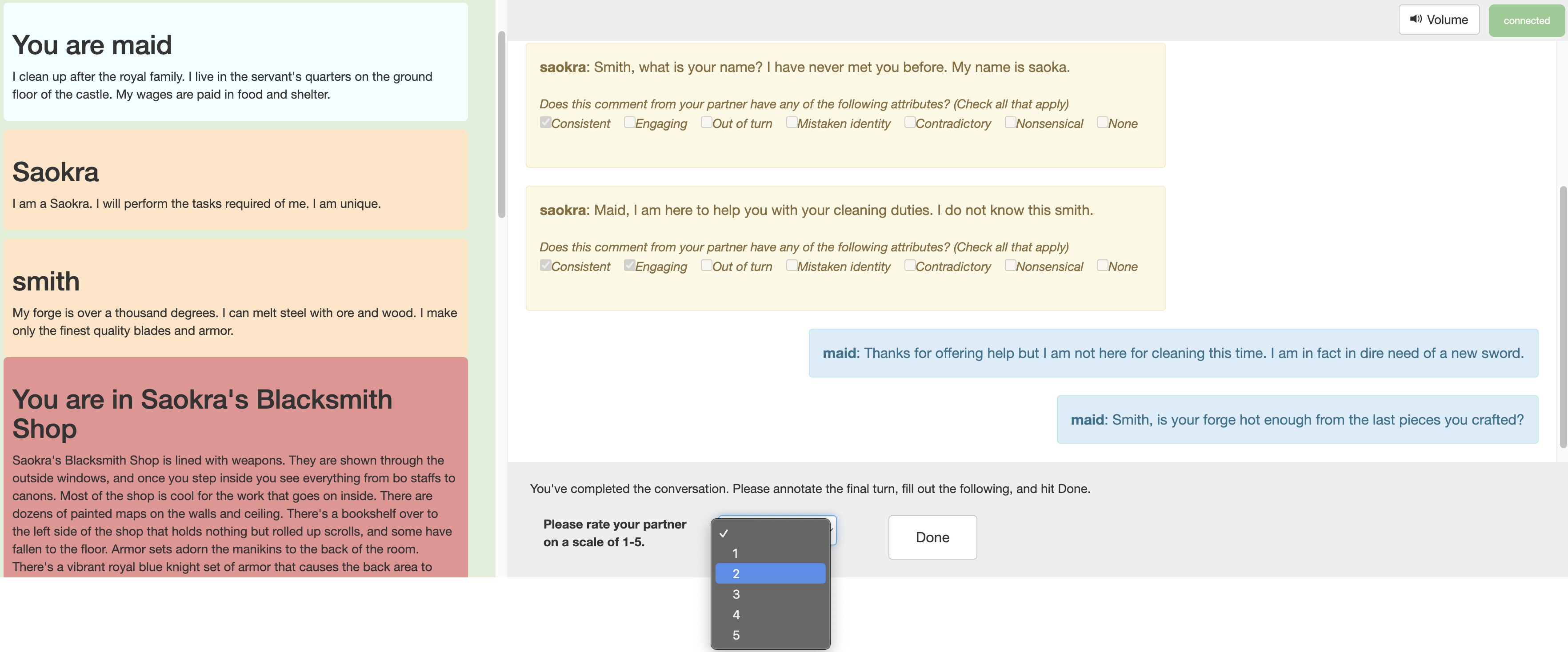}\\
    \caption{Screenshot sample of the task interface used by crowdsourcers for evaluating the models.}
    \label{fig:modelchat_screenshots}
\end{figure*}

This task utilizes Mephisto again, but we also take advantage of ParlAI agents, which build the core of our conversational model (AI agent).
These agents are built on the premise of trun-based conversations and run on a cycle of \emph{observe} and \emph{act}.
We fit our framework of multi-party asynchronous chat into this by running an agent act at the beginning of each round of conversation.
This act first uses the turn-taking model (random or trained) to choose the next speaker.
If the next speaker is determined to be the human, it passes the control to the human and waits for a response,
otherwise it blocks the human from sending a message and meanwhile uses the utterance generation model and produces a response.
When utilizing the utterance generation models we craft the context such that it matches the original format from its training.
In both cases, the response is shown to the model (to be taken into account for updating the history of the conversation).
The cycle repeats until collecting a total of 15 message from either side.

On the user interface side, every time that there is an utterance from the bot,
the process prompts the human to rate the response by attributes that applies to the message (see \autoref{fig:modelchat_screenshots}).
Note that there is a \emph{None} option to choose when none of the selected categories apply.
Also, at the end of the conversation, we ask about the overall quality of the conversation from the human annotators---this is in fact the state that the screenshot in \autoref{fig:modelchat_screenshots} is at.

\end{document}